\documentclass[sigconf]{acmart}

\usepackage{array}
\usepackage{gensymb}
\usepackage{multirow}
\usepackage{lipsum}

\DeclareMathOperator*{\argmin}{arg\,min}

\usepackage{graphics}
\usepackage{array}
\usepackage{multirow}
\usepackage{wrapfig}
\usepackage{hhline}
\usepackage{pifont}%
\newcolumntype{K}[1]{>{\centering\arraybackslash}p{#1}}

\usepackage{duckuments}
\usepackage{tikzducks}
\usepackage{graphicx}
\usepackage{adjustbox}
\usepackage{pifont}
\usepackage{makecell}
\usepackage{hyperref}

\usepackage{adjustbox}
\usepackage{array}
\usepackage{gensymb}
\usepackage{multirow}
\usepackage{lipsum}
\usepackage[normalem]{ulem}
\useunder{\uline}{\ul}{}

\usepackage{atbegshi}
\newcommand{\placetextbox}[3]{
  \setbox0=\hbox{#3}%
  \AtBeginShipoutNext{\AtBeginShipoutUpperLeft{%
    \put(\dimexpr#1\paperwidth\relax,-\dimexpr#2\paperheight\relax)
    {\vtop{{\null}\makebox[0pt][c]{#3}}}%
  }}%
}

\acmSubmissionID{458}

\AtBeginDocument{%
  \providecommand\BibTeX{{%
    \normalfont B\kern-0.5em{\scshape i\kern-0.25em b}\kern-0.8em\TeX}}}

\copyrightyear{2023}
\acmYear{2023}
\setcopyright{acmlicensed}
\acmConference[SA Conference Papers '23]{SIGGRAPH Asia 2023 Conference Papers}{December 12--15, 2023}{Sydney, NSW, Australia}
\acmBooktitle{SIGGRAPH Asia 2023 Conference Papers (SA Conference Papers '23), December 12--15, 2023, Sydney, NSW, Australia}
\acmPrice{15.00}
\acmDOI{10.1145/3610548.3618178}
\acmISBN{979-8-4007-0315-7/23/12}

\usepackage{graphics}
\usepackage{array}
\usepackage{multirow}
\usepackage{wrapfig}

\usepackage{graphicx}
\usepackage{makecell}
\usepackage{hyperref}

\newcommand{\revision}[1]{{#1}}

\usepackage{array}
\usepackage{gensymb}
\usepackage{multirow}
\usepackage{lipsum}

\begin{document}

\title{Intrinsic Harmonization for Illumination-Aware Compositing}

\author{Chris Careaga}
\affiliation{
  \institution{Simon Fraser University}
  \country{Canada}
}

\author{S. Mahdi H. Miangoleh}
\affiliation{
  \institution{Simon Fraser University}
  \country{Canada}
}

\author{Ya\u{g}{\i}z Aksoy}
\affiliation{
  \institution{Simon Fraser University}
  \country{Canada}
}

\renewcommand{\shortauthors}{Careaga et al.}

\begin{abstract}
Despite significant advancements in network-based image harmonization techniques, there still exists a domain disparity between typical training pairs and real-world composites encountered during inference. Most existing methods are trained to reverse global edits made on segmented image regions, which fail to accurately capture the lighting inconsistencies between the foreground and background found in composited images. In this work, we introduce a self-supervised illumination harmonization approach formulated in the intrinsic image domain. First, we estimate a simple global lighting model from mid-level vision representations to generate a rough shading for the foreground region. A network then refines this inferred shading to generate a harmonious re-shading that aligns with the background scene. In order to match the color appearance of the foreground and background, we utilize ideas from prior harmonization approaches to perform parameterized image edits in the albedo domain. To validate the effectiveness of our approach, we present results from challenging real-world composites and conduct a user study to objectively measure the enhanced realism achieved compared to state-of-the-art harmonization methods.\begin{figure*}
    \centering
    \includegraphics[width=\linewidth]{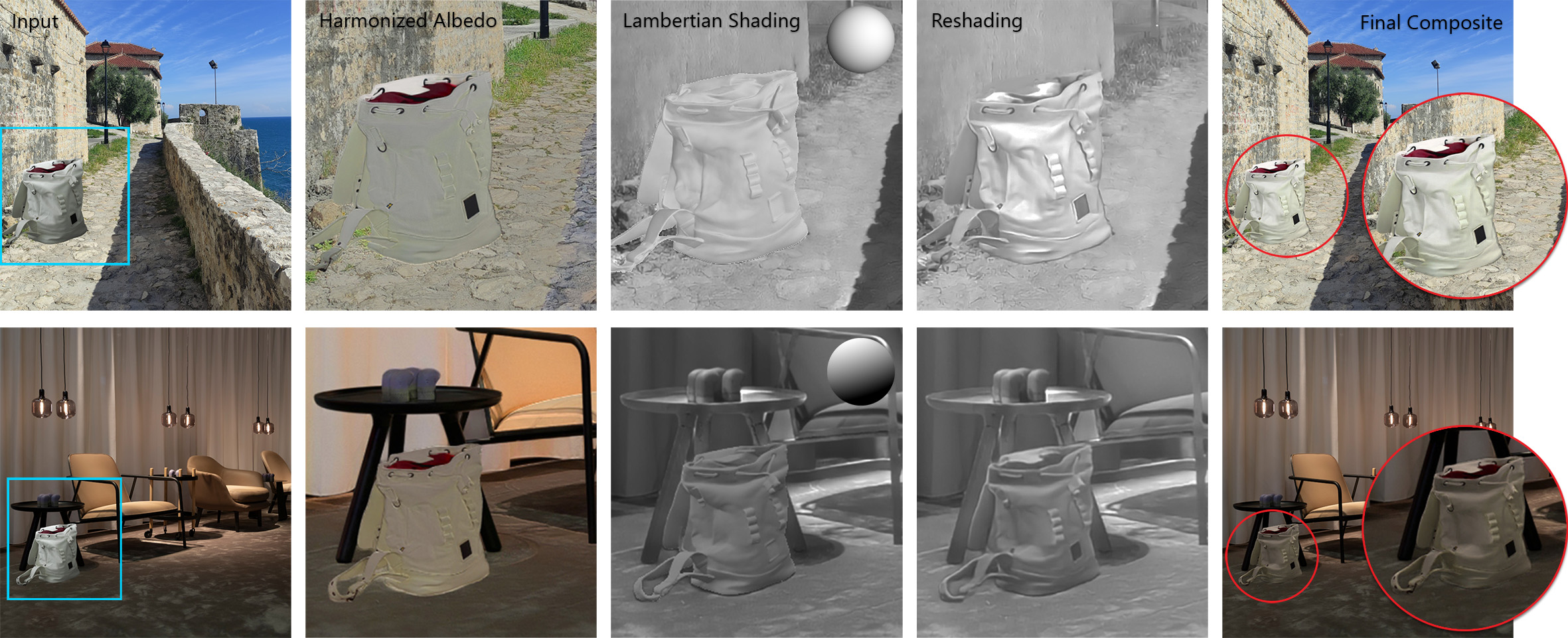}
    \caption{Our method is able to relight foreground regions under varying background lighting conditions. We estimate a lighting model of the background to render a novel shading that is refined by our re-shading network. By harmonizing the albedo, we are also able to represent illumination color. Image credit: Unsplash users Yuliya Yevseyeva, Jean-Philippe Delberghe and \_k8\_}
    \label{fig:teaser2}
\end{figure*}
\end{abstract}

\ccsdesc[500]{Computing methodologies~Image manipulation}

\keywords{image compositing, object relighting, intrinsic decomposition, self-supervised learning, image harmonization}

\begin{teaserfigure}
  \includegraphics[width=\linewidth]{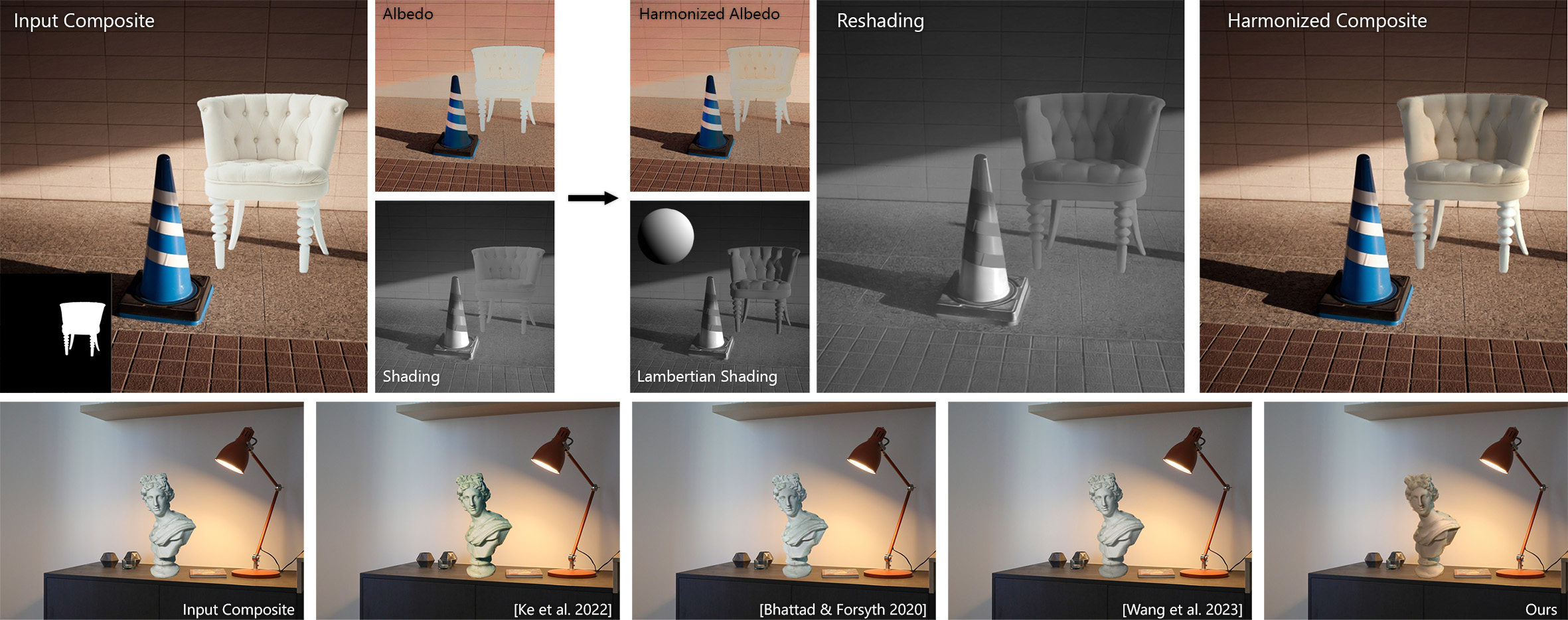}
  \caption{We propose an illumination-aware image harmonization approach for in-the-wild imagery. Our method is formulated in the intrinsic image domain. We use off-the-shelf networks to generate albedo, shading and surface normals for the input composite and background image. We first harmonize the albedo of the background and foreground by predicting image editing parameters. Using normals and shading we estimate a simple lighting model for the background illumination. With this lighting model, we render Lambertian shading for the foreground and refine it using a network trained on segmentation datasets via self-supervision. When compared to prior works we are the only method that is capable of modeling realistic lighting effects. Image credit:  Unsplash users mak\_jp, Daniil Silantev, Jean-Philippe Delberghe and Luo Dan}
  \label{fig:teaser}
\end{teaserfigure}

\maketitle

\placetextbox{0.14}{0.03}{\includegraphics[width=4cm]{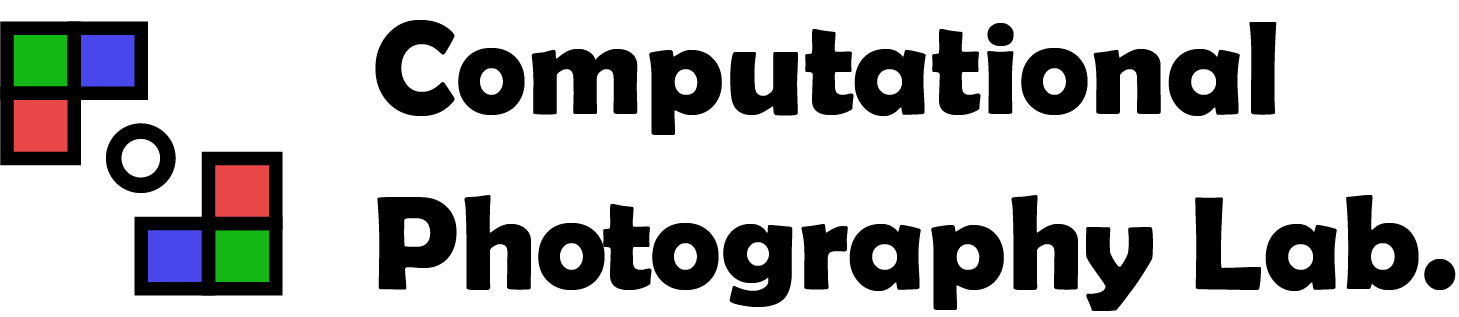}}
\placetextbox{0.85}{0.03}{Find the project web page here:}
\placetextbox{0.85}{0.045}{\textcolor{purple}{\url{https://yaksoy.github.io/intrinsicCompositing/}}}

\section{Introduction}
\label{sec:intro}

Compositing, or inserting an object onto a novel background, is an important image editing task that requires the object to naturally blend in the new environment. 
This requires the appearance of the inserted object to be readjusted to fit the background in a process called image harmonization.
For a realistic composite, the harmonized object should match the color content as well as the illumination present in the background.

Image harmonization has been widely studied in the literature as an image editing problem where self-supervised methods are used to estimate a set of color and tone adjustment operations to match the color contents of the object with the background.
Relighting the object to match the illumination in the novel environment, however, has been largely neglected in the harmonization literature. 
This comes from the difficulty of realistically relighting an object in-the-wild, which requires an accurate estimation of the illuminating environment as well as a detailed and accurate geometry.
As a result, most relighting methods in the literature focus on a specific domain such as portraits~\cite{pandey2021total, yeh2022learning}. 
Restricting the domain of relighting makes it possible to generate training data in controlled environments for image-to-image relighting. 
However, this makes relighting methods in the literature inapplicable to the general problem of image compositing.

In this work, we model the image harmonization problem in the intrinsic domain. 
Intrinsic image decomposition is a fundamental mid-level vision problem that represents an image as the product of the reflectance of the materials and the effect of illumination in the scene:
\begin{equation}
    I = S \cdot A,
\end{equation}
where $S$ and $A$ represent the shading and albedo, respectively. 
By isolating the scene colors contained in the albedo from shading, the intrinsic representation allows us to divide the image harmonization problem into two: color harmonization and relighting. 
In our harmonization pipeline, we first harmonize the color content of the foreground to match the background in the albedo space by borrowing ideas from the rich color harmonization literature. 
We then turn our focus onto the challenging relighting problem in the shading domain. 

Our aim is to generate a new shading for the composited object that reflects the new illumination environment. 
For this purpose, we generate an initial estimation of the shading using a simple Lambertian shading model and surface normals estimated for the background and the inserted object. 
We first estimate the illumination environment of the background using a simplified parametric illumination model. 
We then generate the Lambertian shading for the inserted object and composite this shading map onto the original shading of the background, representing a starting point for our re-shading network. 
Together with the RGB composite, we use the initial shading as input and train our network to generate a realistic new shading for the object. 
This allows us to train our re-shading network in a self-supervised manner using segmentation datasets. 

We show that by dividing the harmonization problem into two and formulating a self-supervised relighting method, realistic composite images can be generated where the inserted object not only reflects the color content of the image but also matches the illumination present in the background.
As shown in Figures \ref{fig:teaser} and \ref{fig:teaser2}, this allows us to generate much more realistic composite images when compared to prior work on image harmonization. 
Through qualitative examples and a subjective evaluation, we demonstrate the performance of our method in challenging scenarios. %

\section{Related Work}
\label{sec:related}

\paragraph{Image harmonization}\quad
Previously proposed image harmonization methods have predominantly been trained in a self-supervised manner in order to undo various image edits performed on a specific region of a natural image \cite{xue2022dccf, cong2020dovenet, ke2022harmonizer, cong2022high}. These edits only represent image-level differences (brightness, saturation, hue, etc.) between the harmonized ground truth and the unharmonized input. This creates a gap between training and inference as real-world composites require more complex operations to be properly harmonized. 
\revision{Various} approaches have been proposed to model the image harmonization problem with more realistic assumptions, including illumination harmonization. \revision{\citet{liao2019approximate} propose a perceptually-inspired shading model to relight image segments. Their approach first estimates a simple geometric model that can be reshaded. They then utilize low-level algorithms to estimate shading effects that are not represented by the rough geometry. Similarly, our approach proposes to utilize estimated geometry that can be reshaded, but replaces the approximate shading model with a data-driven shading refinement process that can be trained via self-supervision.} The method of \citet{guo2021intrinsic, guo2022transformer} formulates an approach based on the intrinsic image domain to jointly separate and harmonize albedo and shading. Their method often fails to generate novel shading due to a lack of proper training data and learning to harmonize and perform intrinsic decomposition with one network. \citet{hu2021neursf} develop a generated dataset and method to relight humans in outdoor scenes. Their method focuses on a specific use case and requires ground-truth geometry as input, making it difficult to use in the wild. Similarly, \citet{bao2022deep} propose a dataset and method to perform illumination harmonization, but they mainly focus on relighting objects on a flat ground plane in outdoor scenes. The work of \citet{bhattad2020cut} formulates an intrinsic approach that re-shades a masked object using Deep Image Prior. Their method fails to generate realistic shading estimations and requires multiple minutes of run time for a single image. \citet{wang2023semi} propose a simpler method of modeling illumination changes in image harmonization, but their method can only locally modulate color-based editing of the foreground region. Given this limitation and lack of explicit reasoning about albedo, they are not able to fully harmonize composites with major differences in lighting between foreground and background.

Our method, on the other hand, borrows ideas from typical image harmonization approaches while also fully modeling the illumination mismatch present in real-world composite images. Using our self-supervised training approach that allows us to employ large-scale segmentation datasets, we are able to perform realistic relighting of foreground regions that match the background lighting environment for in-the-wild composites.

\revision{\paragraph{Object insertion.}\quad}
\revision{Rather than attempting to composite image segments into novel scenes, the task of object insertion attempts to insert a 3D object into a 2D photograph of a scene. This is typically accomplished by \emph{inverse rendering} 2D scenes to infer characteristics of the 3D scene they depict. \citet{moreno2010compositing} proposed an algorithm to recover lighting information from an image segment. They use their lighting information to relight recovered 3D geometry of an object making it appear as if it belongs in the scene. Later works take this idea further, recovering a 3D representation of a scene complete with lighting that can then be used to re-render a 3D object using a typical rendering pipeline \cite{karsch2011rendering, karsch2014automatic}. With the advent of deep learning, inverse rendering techniques have become data-driven, leveraging large-scale synthetic datasets of indoor scenes to train networks for estimating scene intrinsics \cite{li2020inverse, li2022physically}. Our method, on the other hand, does not assume a 3D object as input and instead works completely in the 2D image domain. Additionally, our method is able to work both indoors and outdoors while not requiring large-scale datasets for inferring scene characteristics and instead makes use of off-the-shelf networks for mid-level vision tasks.}

\paragraph{Image relighting.}\quad
Given the complexity of image relighting, prior methods mainly focus on a specific use case such as portraits \cite{pandey2021total, yeh2022learning} or outdoor structures \cite{griffiths2022outcast}. \revision{These methods rely on large-scale, difficult-to-obtain datasets, or multi-view scenes \cite{philip2019multi, philip2021free, nicolet2020repurposing}, in order to achieve realistic results}.

Additionally, if the target illumination is not provided, as is the case in image harmonization, lighting must be predicted. Generating estimations of lighting configuration is a difficult task also relying on hard-to-capture datasets \revision{\cite{zhang2019all, garon2019fast}}. By leveraging off-the-shelf mid-level estimations, our method learns to relight the foreground from a single image, in the wild, without requiring ground-truth source and target illumination examples. Additionally, we utilize a simple lighting estimation formulation to guide our re-shading network, while still generating realistic shading estimations.

\begin{figure*}
    \centering
    \includegraphics[width=\linewidth]{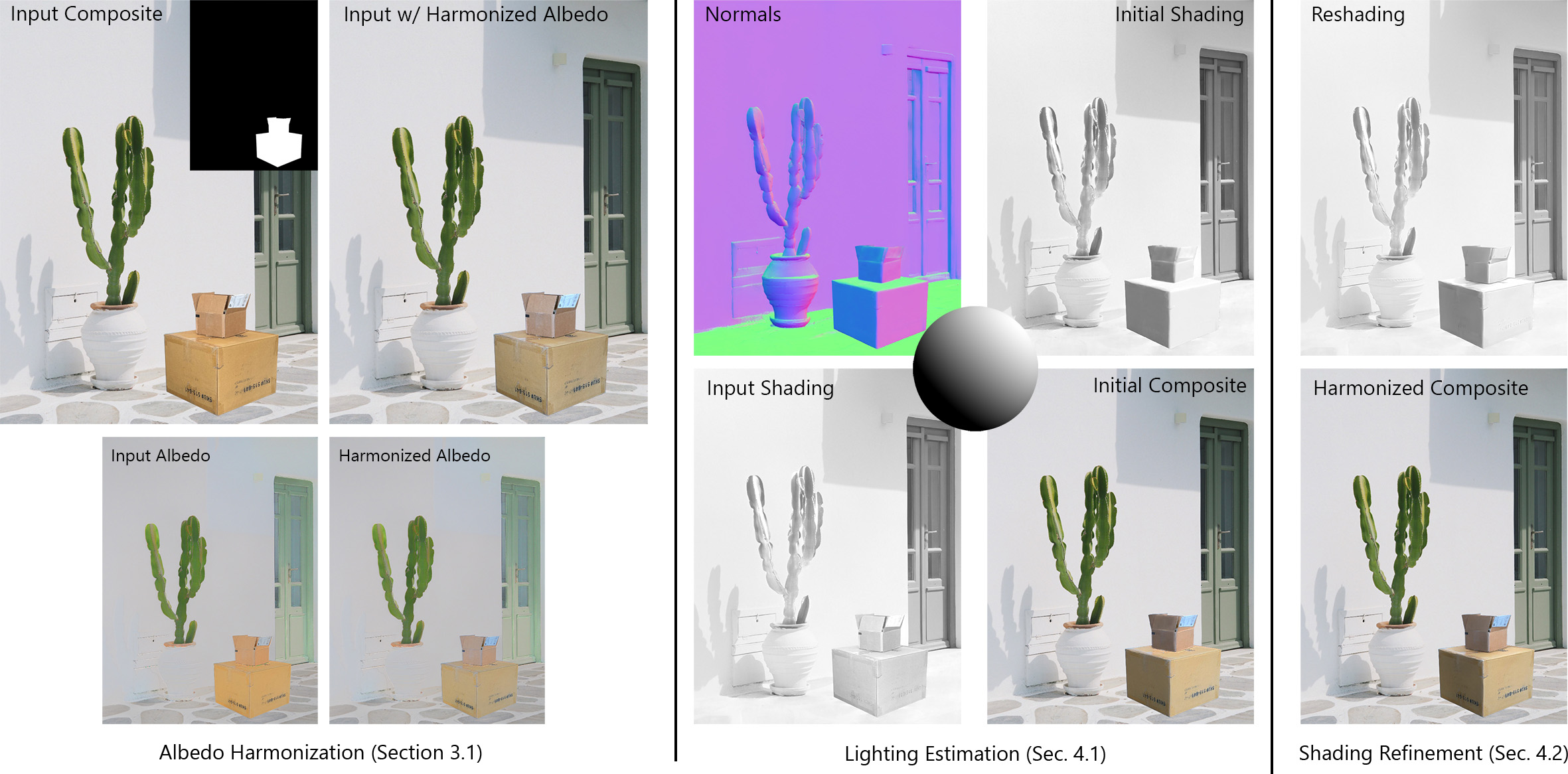}
    \caption{Our pipeline consists of three main parts. We first perform harmonization on the estimated albedo of the scene (Section~\ref{sec:overview:albedo}). We next use estimated normals and shading of the background to optimize the parameters of a simple lighting model (Section~\ref{sec:reshading:lighting}). Finally, we use the lighting model to render a Lambertian shading of the foreground region. We refine this Lambertian shading with a network to generate a final realistic shading layer that we combine with our harmonized albedo to create our final result (Section~\ref{sec:reshading:refinement}). Image credit: Unsplash user Chlo\'{e} Chavanon}
    \label{fig:pipeline}
\end{figure*}

\section{Intrinsic Harmonization}
\label{sec:overview}

\begin{figure*}[t]
    \centering
    \includegraphics[width=\linewidth]{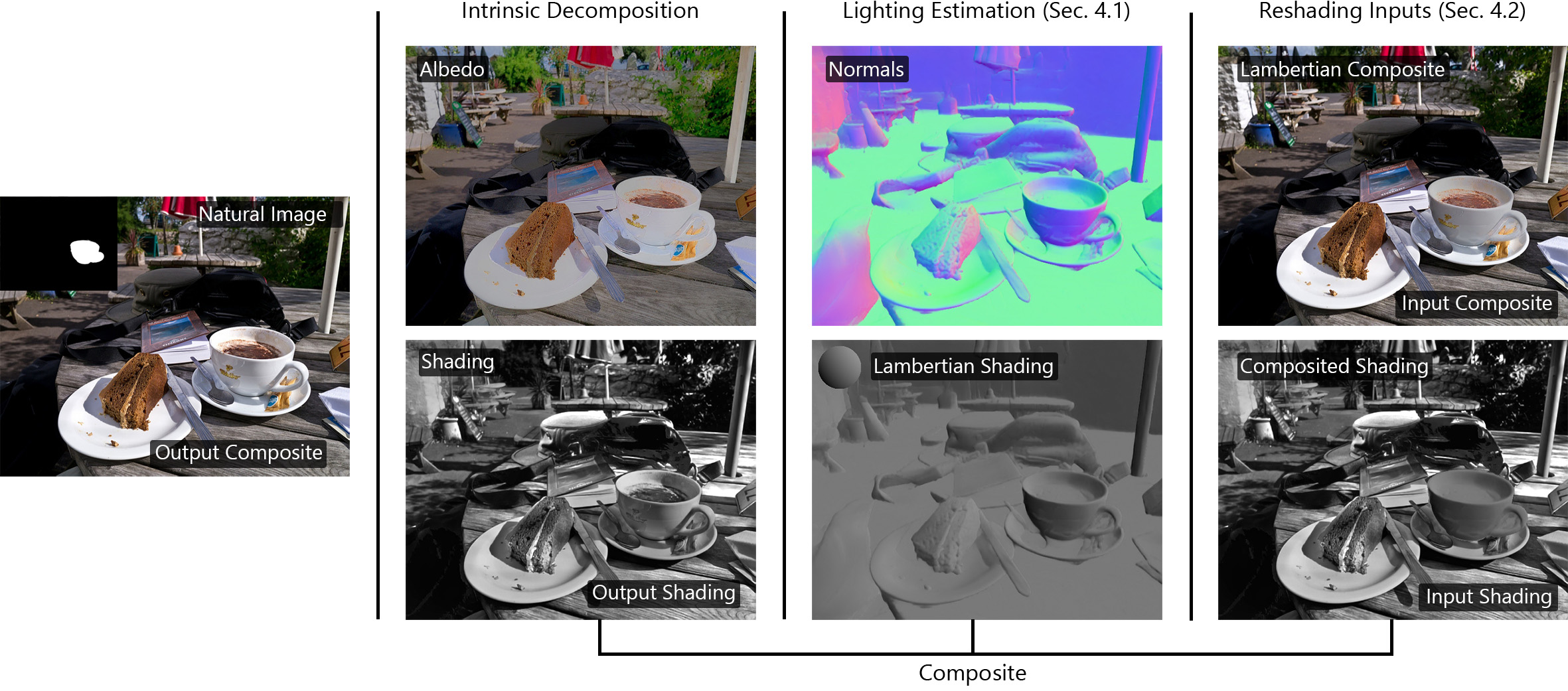}
    \caption{To train our re-shading network (Section~\ref{sec:reshading:refinement}) we propose a self-supervised data generation strategy. We use image segmentation datasets to provide masked image regions for diverse scenes. We first perform intrinsic decomposition on the image. We use our lighting estimation method (Section~\ref{sec:reshading:lighting}) to generate lighting parameters for the non-masked region. We render Lambertian for the foreground region and composite onto the background shading. This shading is used, along with the original albedo, to generate a new composite with Lambertian shading in the masked region. Our model is trained to reverse this process, learning to map the Lambertian shading to accurate and realistic shading. Image credit: Flickr user shirokazan}
    \label{fig:self_supervision}
\end{figure*}

Our approach makes use of multiple off-the-shelf methods to generate mid-level representations. We denote the background scene and the foreground object to be composited as $I_b$ and $I_f$, respectively. 
We start by generating the intrinsic decomposition of both images using the method by \citet{chrisIntrinsic}:
\begin{equation}
    I^f = S^f \cdot A^f, 
    \quad \quad I^b = S^b \cdot A^b.
    \label{eq:intrinsic}
\end{equation}
Where $S$ and $A$ denote the single channel shading and the RGB albedo, respectively. 
We define the compositing in the albedo and the shading separately:
\begin{equation}
    A^c = \alpha A^f + (1-\alpha) A^b, 
    \quad S^c = \alpha S^f + (1-\alpha) S^b, 
    \quad I^c = A^c \cdot S^c.
\end{equation}
where $\alpha$ represents the foreground mask, and $I^c$, $A^c$, and $S^c$ represent the composite image and its intrinsic components. \revision{Our approach uses linear RGB when performing any albedo and shading operations as Equation \ref{eq:intrinsic} assumes linear RGB values. When given a standard RGB image as input we reverse the gamma-correction process using a gamma value of 2.2.}
This separation of the compositing problem allows us to perform color and illumination harmonization in two separate steps. 
We will first give an overview of our albedo harmonization, and then detail our self-supervised approach to generate the new foreground shading in Section~\ref{sec:reshading}. Examples of our harmonized albedo can be seen in Figures \ref{fig:teaser2} and \ref{fig:pipeline}.

\subsection{Albedo Harmonization}
\label{sec:overview:albedo}
The color content of a scene is represented in the albedo, therefore we design a simple color-based harmonization method to adjust the foreground albedo to harmonize well with the colors in the background albedo. Similar to prior parameter-based harmonization approaches \cite{wang2023semi, ke2022harmonizer}, we aim to find a set of editing parameters that control common image editing operations such as changing the exposure, saturation, color curve, and white balance. We achieve this through a self-supervised setup using segmentation datasets. For a given segment, we apply a random set of operations to the object to create a mismatch between the simulated \emph{composited} albedo and the original image. 
We then train a network to estimate the editing parameters that will re-create the original appearance of the object in the scene, $A^c$. 
For this purpose, we utilize the editing network proposed by \citet{miangolehrealistic}. We train our network with the mean squared error defined on the RGB albedo using the MS COCO~\cite{lin2014microsoft} and Davis \cite{perazzi2016davis} datasets for 100 epochs. We provide a detailed description of our setup in the supplementary material.

\section{Foreground Re-shading}
\label{sec:reshading}

Harmonizing the illumination of the foreground region in a composite image can be posed as a specific case of image relighting. 
Achieving physically-accurate relighting would require an accurate HDRI illumination environment together with a highly detailed geometry to physically re-render the object. 
However, the creation of a relighting dataset for data-driven relighting is highly challenging and hard to scale due to the controlled setup such a dataset requires.
Instead of training a network for image-to-image relighting, in order to create a novel foreground shading that matches the background environment, we develop a simple Lambertian shading model with a parametric illumination representation. 
We then train a network to refine our Lambertian shading in order to generate a realistic shading map for the final composite image.
We show that by defining the relighting problem as the refinement of the Lambertian shading, we can train our network in a self-supervised manner using standard segmentation datasets.

\subsection{Parametric Illumination Model}
\label{sec:reshading:lighting}
\begin{figure*}[t]
    \centering
    \includegraphics[width=\linewidth]{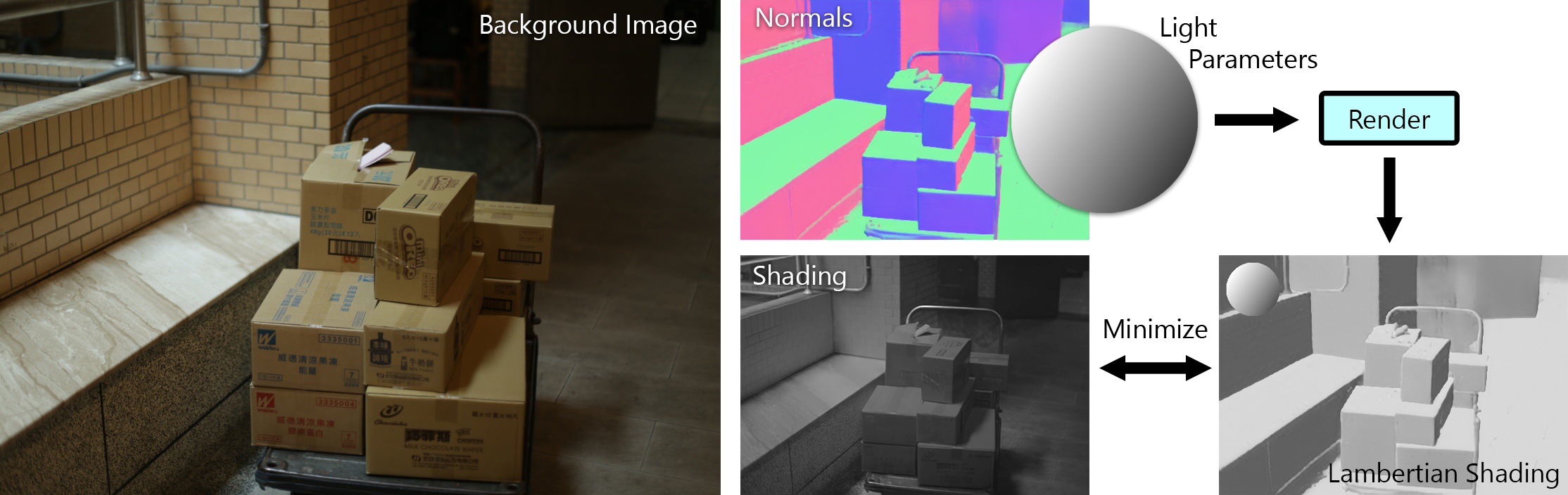}
    \caption{Using estimated normals and shading, we optimize lighting parameters such that the rendered Lambertian shading best matches the estimated shading of the scene. We use this lighting representation to infer the general shading of an object in the scene. Image credit: Unsplash user ej1209\_}
    \label{fig:light_estimation}
\end{figure*}

We first start our re-shading pipeline by estimating a simple illumination model for the background scene using the Lambertian shading model. 
We define our illumination model as a combination of a directional light source represented by the vector $\vec{l} \in \mathbb{R}^3$ and a constant ambient illumination $c$. 
The Lambertian shading model represents the shading of a scene using the surface normals and the illumination environment. Using our simple illumination model, the Lambertian shading for the background $\tilde{S}^b$ is defined as:
\begin{equation}
    \tilde{S}^b_i = \vec{n}^b_i \cdot \vec{l} + c,
\end{equation}
where $\vec{n}^b_i$ represents the surface normal at pixel $i$ estimated using an off-the-shelf method~\cite{eftekhar2021omnidata}. 
We estimate the parameters of our lighting model with a least squares optimization:
\begin{equation}
    (\vec{l}, c) = \argmin_{\vec{v}, x} \sum_{i} \left( S^b_i - \vec{n}^b_i \cdot \vec{v} + x \right)^2.
\end{equation}
In other words, we search for the setting of $\vec{l}$ and $c$ such that the rendered Lambertian shading best reconstructs the estimated shading of the background scene.  
In order to ensure the lighting model parameters take on plausible values, we constrain the values of \revision{$\vec{l}$} and $c$ to be positive. This keeps \revision{$\vec{l}$} in the outward-facing hemisphere. We solve this minimization problem using gradient-based optimization with the Adam optimizer \cite{kingma2014adam}. We show an example of our lighting estimation pipeline in Figure~\ref{fig:light_estimation}.

We use the estimated illumination model to generate the Lambertian shading for the foreground object:
\begin{equation}
    \tilde{S}^f_i = \vec{n}^f_i \cdot \vec{l} + c,
\end{equation}
which represents the starting point for our shading refinement process. We composite $\tilde{S}^f$ onto $S^b$ to generate the initial composite shading $\tilde{S}^c$, which we use as input for our shading refinement.

\subsection{Shading Refinement}
\label{sec:reshading:refinement}
We define the re-shading problem as the refinement of $\tilde{S}^c$ into a realistic shading map $S^c$ such that when multiplied with the albedo, reconstructs the illumination-harmonized composite image. 
This definition allows us to train our re-shading network in a self-supervised fashion. 
We generate our input - ground-truth pairs using real images with a segmentation mask, representing the composited region. 
We then estimate the Lambertian shading of the object using our parametric model, which is used as the input to our network. 
The ground-truth shading is then defined as the shading estimated for the original image $S^c$. 
The input to our network is the foreground mask, the RGB composite image with Lambertian shading, $A^c \cdot \tilde{S}^c$, the Lambertian shading $\tilde{S}^c$, as well as the surface normals and a depth map generated using methods by \citet{eftekhar2021omnidata} and \citet{miangoleh2021boosting} to provide our network with geometric context. We channel-wise concatenate the inputs for a $h \times w \times 9$ input, $h$ and $w$ representing the height and width of the image, respectively. An overview of the self-supervised data generation process can be seen in Figure~\ref{fig:self_supervision}. We supervise our network using losses defined on the shading and the final reconstructed RGB image, $I^c = A^c \cdot S^c$. 
We use the mean squared error:
\begin{equation}
    \mathcal{L}_s = MSE(\hat{S}^c, S^c), \quad \mathcal{L}_i = MSE(\hat{I}^c, I^c),
\end{equation}
where $\hat{S}^c$ is the estimated shading and $\hat{I}^c = A^c \cdot \hat{S}^c$ is the estimated composite image. 
We also use the multi-scale gradient loss \cite{li2018mega} as an edge-aware smoothness loss:
\begin{equation}
    \mathcal{L}_{sg} = \sum_m MSE(\nabla \hat{S}^{c,m}, \nabla S^{c,m}),
    \quad
    \mathcal{L}_{ig} = \sum_m MSE(\nabla \hat{I}^{c,m}, \nabla I^{c,m}),
\end{equation}
where $\nabla S^{c,m}$ represents the gradient of $S^c$ at scale $m$. 
We combine these losses to compute our final loss used during training:
\begin{equation}
    \mathcal{L} = \mathcal{L}_s + \mathcal{L}_i + \mathcal{L}_{sg} + \mathcal{L}_{ig}
\end{equation}
using unit weights.

\subsection{Network Architecture and Training}

We combine 3 datasets to train our re-shading network: the COCO Dataset \cite{lin2014microsoft}, a 50,000 image subset of the SA-1B Dataset \cite{kirillov2023segany}, and the Multi-Illumination Dataset (MID) \cite{murmann2019multi}. For COCO and SA-1B, we utilize the provided segmentation masks and sample foreground segments that are sufficiently large. 
For MID, we use the segments provided in the dataset and use the provided multiple illuminations for dataset augmentation. We use the encoder-decoder architecture used by \citet{ranftl2020towards} which consists of a ResNext101 \cite{xie2016aggregated} encoder and a RefineNet \cite{lin2017refinenet} decoder. We train our network using the Adam optimizer with a learning rate of $1 \times 10^{-5}$ for 2 million iterations.

Our definition of the relighting problem as shading refinement allows us to use generic segmentation datasets for self-supervised training. As these datasets provide diverse training data for both indoor and outdoor conditions, we are able to perform illumination harmonization, i.e. object relighting in-context, in-the-wild.

\begin{figure*}
    \centering
    \includegraphics[width=\linewidth]{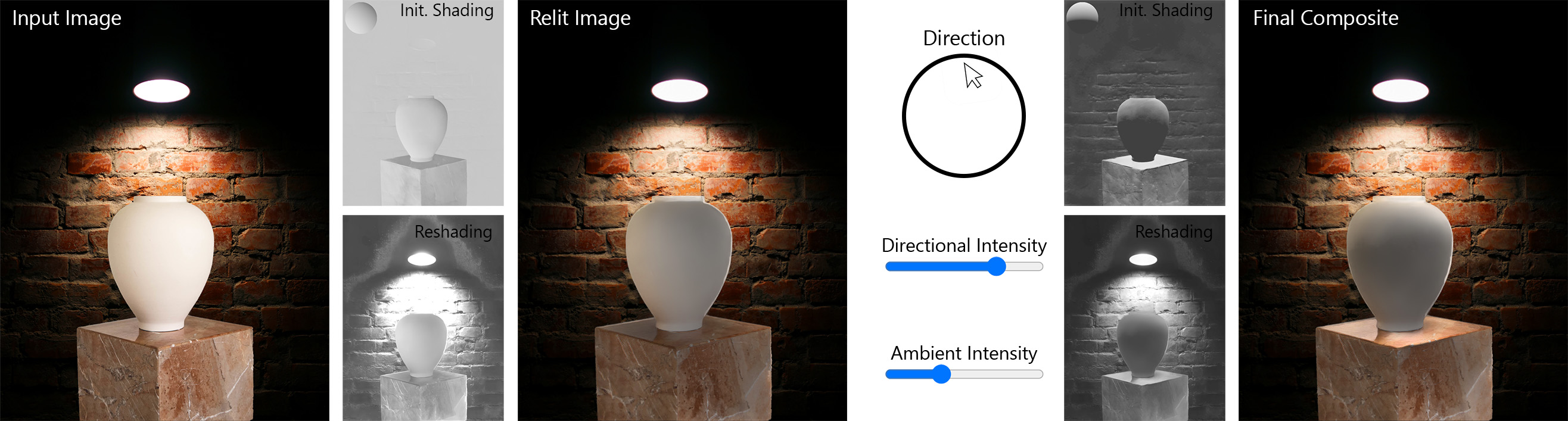}
    \caption{While our simple lighting model can fail when our assumptions are broken, due to its small set of interpretable parameters it can be easily edited by a human to achieve the desired result. This aspect of our approach could be incorporated into a GUI to interactively harmonize an image. Image credit: Unsplash users Mathilde Langevin and De an Sun}
    \label{fig:interactive_edit}
\end{figure*}
\section{Experimental Evaluation}
\label{sec:experiments}

To evaluate the realism of our generated composites we compare against various state-of-the-art image harmonization methods. We focus our evaluation on methods that also attempt to model illumination harmonization. We perform a user study comparing multiple methods on difficult composite images. Additionally, we provide qualitative comparisons of the scenes used in our user study.

\begin{table}[t]
\caption{Bradley-Terry \cite{bradley1952rank} ranking scores of naive composites and state-of-the-art methods compared to our models. While our model without reshading performs comparably to \citet{wang2023semi}'s model, our full model with reshading achieves the highest score with a significant gap.  }
{
\begin{tabular}{l c}
    Method & B-T Score $\uparrow$ \\
    \hline
    Naive Composite & 0.0933 \\
    \citet{bhattad2020cut} & 0.0893 \\
    \citet{ke2022harmonizer} & 0.1727 \\
    \citet{wang2023semi} & 0.2078 \\
    \hline
    Ours (w/o reshading) & 0.1906 \\
    Ours (full) & \textbf{0.2485} \\
\end{tabular}
}
\label{tab:user_study}
\end{table}

\subsection{User Study}

In order to concretely evaluate the effectiveness of our harmonization, we perform a qualitative user study. We model our user study after that of \citet{wang2023semi} and perform a \textit{two alternatives forced choice} survey. We compare our method to three state-of-the-art methods. The method of \citet{bhattad2020cut} also performs illumination harmonization in the intrinsic domain. \citet{wang2023semi} utilize a gain map to modulate image edits allowing them to model non-global edits. The work of \citet{ke2022harmonizer} proposes a parametric harmonization approach for high-definition images. Each of these methods can produce results at high-resolution making them suitable for qualitative comparison against our method. 

We generate 50 composites using free-to-use images from Unsplash. We aim to create difficult examples with a mismatch in color and illumination between the foreground and background regions. Examples of our composited images are shown in Figure~\ref{fig:comp:all:study}. For each composite, our method is compared to the naive input composite, three prior works, and our method without shading harmonization. We generate image pairs consisting of our method and each of the comparison methods for each image, resulting in $5 \times 50 = 250$ pairs. Each participant is given context about image compositing and told to \textit{``determine which image has the foreground object better matching the background environment''}. They are then shown \revision{a random set of 50 pairs without duplicate photographs}. We crowd-sourced our survey via word of mouth on social media. We obtained responses from 70 subjects, resulting in 3500 total comparisons. We follow prior harmonization works \cite{wang2023semi, cong2020dovenet} and analyze our responses using the Bradley-Terry model \cite{bradley1952rank} to generate global ranking scores for each method. The results of our user study are shown in Table \ref{tab:user_study}. We observe that our method is preferred over all other methods. Additionally, we observe that our model is preferred significantly more when the illumination is harmonized, showing that illumination is an important aspect of realism when it comes to in-the-wild composites.

\subsection{Qualitative Comparisons}

\begin{figure*}[t]
    \centering
    \includegraphics[width=\linewidth]{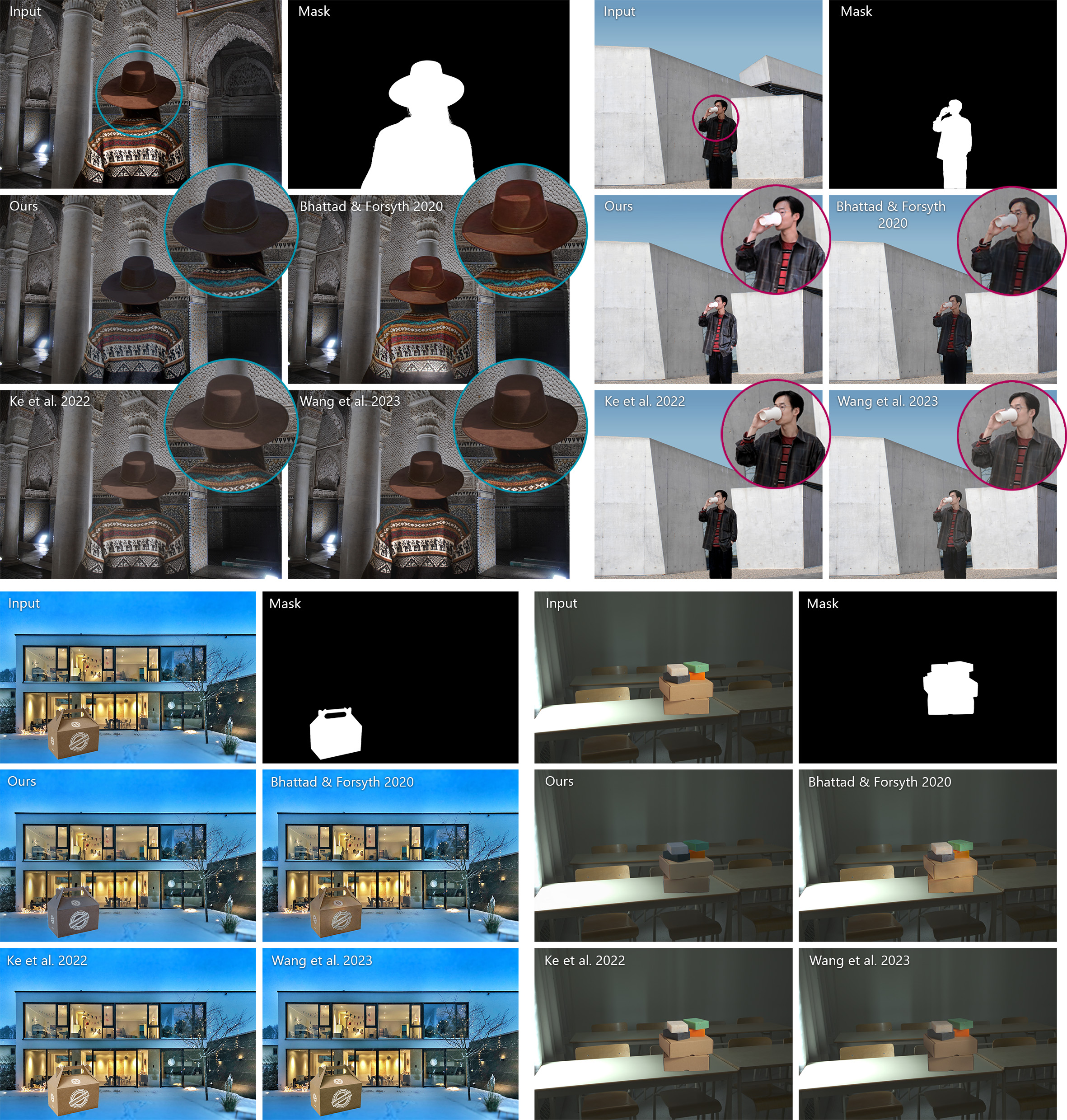}
    \caption{We show comparisons to prior works on difficult composites. Prior methods are not able to fully model the relighting of the foreground, and therefore generate composites with the lighting of the original foreground image. This mismatch results in unrealistic composite images as shown by our user study. Image credit: Unsplash users Stephan Bechert, Grupo Seripafer, JR Harris, victor\_g, Jeison Higuita, Miguel Constantin Montes, Jason An and The Laval Indoor Spatially Varying HDR Dataset \cite{garon2019fast}}
    \label{fig:comp:all:study}
\end{figure*}

Figure ~\ref{fig:comp:all:study} shows qualitative comparisons on composite images gathered as part of our user study. We show that prior works fail to harmonize the illumination differences between the foreground and background. In the first row, none of the prior works are able to attenuate the shadow on the hat from the foreground's original outdoor environment. In the second row, our method is able to estimate the bright outdoor light shining on the building facade in the background. In the third row, the prior works fail to soften the lighting on the box, leaving the original direct lighting that does not match the blue ambient lighting from the background. In the bottom row, our method is able to dim the overall brightness of the illumination on the boxes and also reflect the direction of the light coming from the window.

\begin{figure*}
    \centering
    \includegraphics[width=\textwidth]{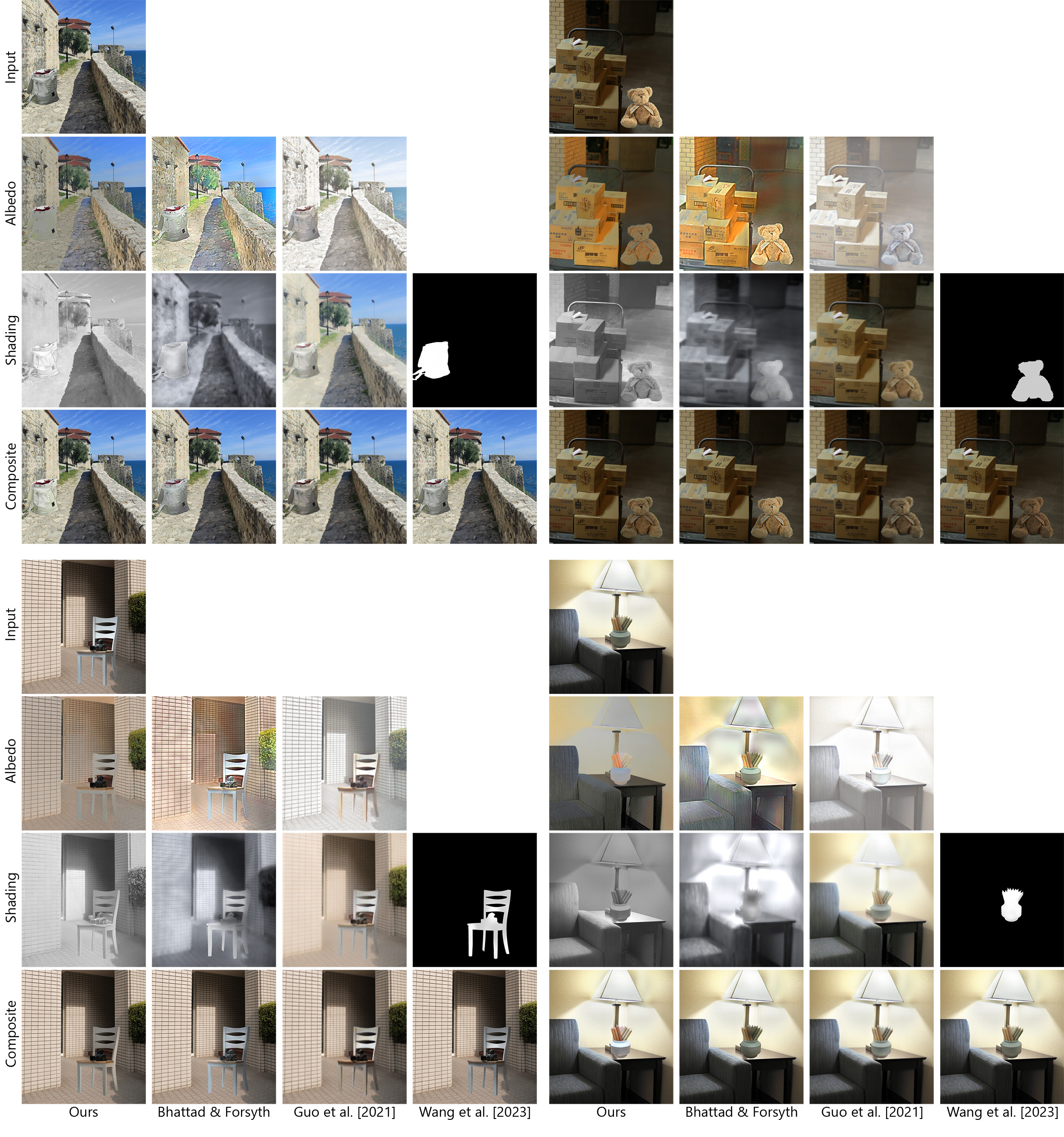}
    \caption{Our method is able to generate physically accurate novel re-shadings under a variety of conditions. Although the works of \citet{guo2021intrinsic} and \citet{bhattad2020cut} attempt to model the illumination in the form of intrinsic components, they fail to generalize and are not able to estimate meaningful representations. The method of \citet{wang2023semi} is able to modulate parametric edits to simulate lighting alteration, their approach doesn't explicitly albedo and shading and therefore cannot generate novel illuminations. Image credit: Unsplash users Laura Ohlman, mak\_jp, Solstice Hannan, Randy Fath and La{\aa}rk Boshoff}
    \label{fig:comp:withcomponents}
\end{figure*}

Figure~\ref{fig:comp:withcomponents} provides a comparison to prior works that also model intrinsics, or perform non-global edits to simulate altered lighting. We include the method of \citet{guo2021intrinsic} in our comparison as they also model the harmonization method in the intrinsic domain. We do not include their results in our user study as they can only estimate at a resolution of $(256 \times 256)$. We observe that prior works struggle to estimate accurate intrinsic representations. The approach of \citet{bhattad2020cut} cannot generate high-frequency details in their shading due to estimating at low resolution, therefore their results typically don't exhibit novel illumination effects. The work of \citet{guo2021intrinsic} attempts to perform intrinsic decomposition and harmonization jointly, but fails to estimate meaningful albedo and shading due to a lack of ground-truth supervision for these quantities. The method of \citet{wang2023semi} predicts a gain map to modulate parametric edits. While their gain map does allow them to model local variations, their method cannot fully relight the foreground as they do not explicitly model albedo. This results in their harmonized foreground maintaining the illumination conditions from its source environment. Our method models the problem with physical accuracy and can therefore predict a detailed and accurate re-shading of the foreground.

\section{Discussion}
\label{sec:discussion}
While we propose a pipeline for performing illumination-aware image harmonization, our re-shading network can also be utilized for interactive relighting. Given our simple lighting model, it is easy to alter the illumination conditions of the Lambertian shading given to our network. As Figure~\ref{fig:teaser2} shows, our re-shading network faithfully generates a realistic shading layer that obeys the conditions of the input Lambertian shading. This aspect of our method is also amenable to interactive image editing applications. As Figure~\ref{fig:interactive_edit} shows, in case our estimated parametric light model is inaccurate, a user can easily alter our lighting model by providing values for the lighting parameters. The user can then yield a more pleasing composite. When combined with our parametric albedo harmonization formulation, a user can have full control over the image harmonization process.

\section{Limitations}
\label{sec:limitations}
Our method relies on multiple off-the-shelf networks to estimate various mid-level vision representations. Although our pipeline is generally robust to small errors in these estimations, an inaccurate intrinsic decomposition or surface normals may lead to inaccuracies in in the lighting model estimation or in the Lambertian shading generated for the foreground. It does, however, mean that if these methods are improved, our model would see a similar gain in performance. Additionally, our lighting model is also rudimentary and cannot represent colorful illumination and hence we rely on the albedo harmonization network to account for any illumination color coming from the background environment. Given the assumptions of the Lambertian shading model, our lighting estimation can fail in difficult scenarios such as multiple colored light sources in the image. Finally, while we are able to relight the foreground object, our method does not model the cast shadows that the object may generate in the new environment. This requires a detailed understanding of the geometry of the background environment, and we believe this is a limitation to address in future work. 

\section{Conclusion}
\label{sec:conclusion}
We introduce a self-supervised illumination harmonization approach formulated in the intrinsic image domain. We show that the intrinsic domain allows us to address two challenges in image harmonization, color and illumination mismatch, separately using dedicated models. We utilize ideas from the prior image harmonization literature to first match the color appearance of the foreground and background albedo by performing parameterized image edits. We then estimate a simple global lighting model from mid-level vision representations to generate Lambertian shading for the foreground region. We train a network via self-supervision to refine this inferred shading into a realistic re-shading that aligns with the background scene. This represents an interesting direction in the relighting literature, which typically models the problem as image-to-image translation trained on specialized datasets. To show the effectiveness of our approach, we conduct a user study on challenging real-world composites to objectively measure the enhanced realism achieved compared to state-of-the-art harmonization methods.

\begin{acks}
The authors would like to thank Elise Saxon for helping with the creation of user study surveys and Ke Wang for promptly providing data and results for their method. We acknowledge the support of the Natural Sciences and Engineering Research Council of Canada (NSERC), [RGPIN-2020-05375].
\end{acks}

\bibliographystyle{ACM-Reference-Format}
\bibliography{references}


\end{document}